# Modeling, Control and Self-sensing of Dielectric Elastomer Soft Actuators: A Review

Yunhua Zhao and Guang Meng

*Abstract*—Dielectric elastomer actuators (DEAs) have garnered extensive attention especially in soft robotic applications over the past few decades owing to the advantages of lightweight, large strain, fast response and high energy density. However, because the DEAs suffer from nonlinear elasticity, inherent viscoelastic creep, hysteresis and vibrational dynamics, the modeling, control and self-sensing of DEAs are challenging, thereby hindering the practical applications of DEAs. In order to address these challenges, numerous studies have been conducted. In this review, various physics-based modeling methods and phenomenological modeling methods for predicting the electromechanical response of DEAs are presented and discussed. Different control methods for DEAs are reviewed, which are classified into open-loop feedforward control, feedback control, feedforward-feedback control and adaptive feedforward control. Physics-based self-sensing methods and data-driven self-sensing methods for reconstructing the DEA displacement without the need for additional sensors are discussed. Finally, the existing problems and new opportunities for the further studies are summarized.

## I. INTRODUCTION

DIELECTRIC elastomer actuators (DEAs) have been hailed as the most promising artificial muscles, because they exhibit inherent flexibility, large actuation strain, high energy density, lightweight and fast response [1], [2]. The unusual combination of excellent performances makes DEAs attractive for many applications, such as soft robotics [3], [4], [5], [6], haptic feedback [7], [8], [9], [10] and vibration isolation [11], [12], [13], [14]. Over the past few decades, the development of DEAs has grown rapidly, but most studies have focused on elastomer/electrode materials and fabrication technique (to improve the actuation performance and life cycle, as well as to decrease the driving voltage) [11], [15], [16], [17], [18], configuration design (for various applications) [19], [20], [21] and quasi-static modeling (to gain an insight into the electromechanical instability and dielectric breakdown) [22], [23], [24]. With leveraging the high-frequency characteristics of DEAs for a wider range of fields, such as pocket pumps [25], soft valves [26], and flapping-wing microrobots [4], [27], the research on dynamic modeling and tracking control of DEAs has been rapidly increasing in recent years. This is because precise control is crucial for DEAs or DEA-driven devices to accomplish the intended tasks with accuracy. In addition, DEAs have an attractive feature—self-sensing, which means that a DEA can function as an actuator and a sensor simultaneously. Thus, in principle, the self-sensing can be utilized in the closed-loop control of DEAs to provide signal feedback without relying on additional sensors, thereby enabling a more compact and cost-effective system. Various self-sensing approaches have been proposed, which determine the DEA output based on the measured electrical quantities. This paper intends to comprehensively review the dynamic modeling, tracking control and self-sensing of DEAs (Fig. 1), and also highlights the existing challenges and opportunities for further studies. It should be pointed out that there is no specific frequency range for the DEA dynamics, as the bandwidth of a DEA varies with different dielectric materials and different electrode materials; the dynamics mentioned here means that the time-dependent and frequent-dependent effects of the actuators' response are taken into account.

The remaining parts of the paper are organized as follows. Firstly, a description of the DEA system is given and the electromechanical behaviors it exhibits are introduced in Section II. Section III reviews various physics-based modeling methods and phenomenological modeling methods for DEAs. Then in Section IV, the control strategies for DEAs are presented and discussed, which are classified into open-loop feedforward control, feedback control, feedforward-feedback control and adaptive feedforward control. Physics-based self-sensing methods and data-driven self-sensing methods are reviewed in Section V. Finally, the summary and outlook are given in Section VI.

## II. DIELECTRIC ELASTOMER ACTUATORS

### A. System Description

Fig. 2(a) shows a schematic of the experimental setup for testing and controlling a DEA. In order to ensure that the DEA can produce an appreciable stroke, a voltage in the kilovolt range applied across the dielectric film is usually required. Thus, a high-voltage amplifier is used to amplify the signal from the data acquisition (DAQ) module and drive the DEA.

This work was supported in part by the State Key Laboratory of Mechanical System and Vibration (Grant No. MSV202616), and in part by the Hebei Provincial Education Department Project (Grant No. QN2026819). (Corresponding author: Yunhua Zhao.)

Yunhua Zhao is with the School of Mechanical Engineering, Hebei University of Technology, Tianjin 300401, China (e-mail: yunhua_zhao@hebut.edu.cn).

Guang Meng is with the State Key Laboratory of Mechanical System and Vibration, School of Mechanical Engineering, Shanghai Jiao Tong University, Shanghai 200240, China (e-mail: gmeng@sjtu.edu.cn).

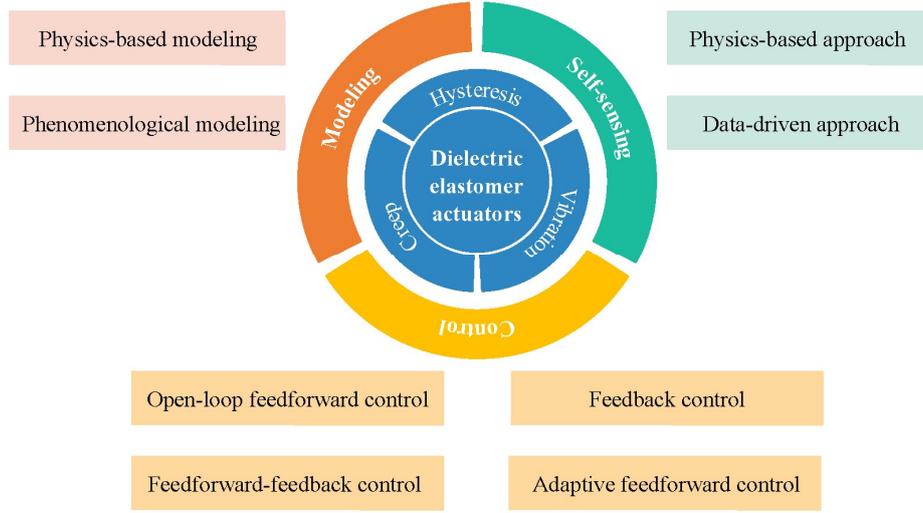

Fig. 1. Overview of modeling, control and self-sensing of DEAs.

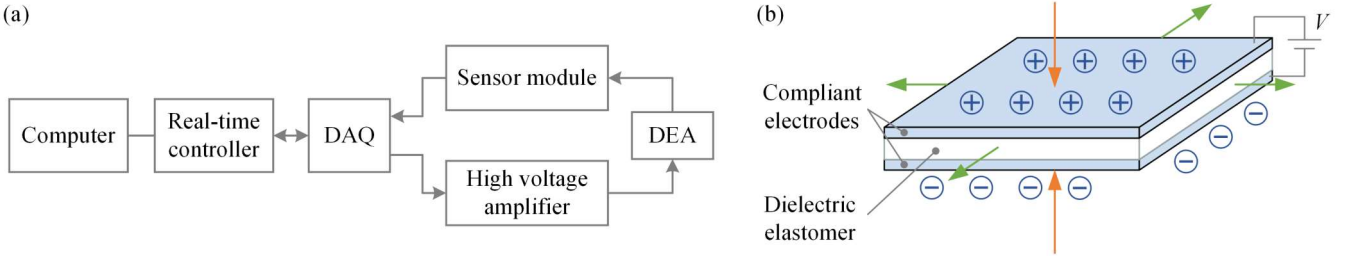

Fig. 2. (a) Block diagram of the experimental setup for testing and controlling a DEA. (b) DE actuation principle.

The DEA's output displacement signal is measured and conditioned in the sensor module. The control voltage input to the DEA is adjusted according to the developed control algorithm running on the real-time controller.

As depicted in Fig. 2(b), the basic element of a DEA consists of an elastomeric polymer film sandwiched between two compliant electrodes. When a voltage is applied across the electrodes, the induced electrostatic pressure results in thickness reduction and in-plane expansion of the dielectric elastomer (DE) film. The simple actuation principle and inherent passive compliance of DEs enable the DEAs to be designed into a variety of different configurations for extensive applications. For example, the elastomer films can be rolled into a scroll, laminated on a flexible substrate, or stretched over a frame [28]. These DEAs are developed by exploiting the area expansion of elastomer films; other configurations, such as multi-layer stacked [29], [30], helical [31] and folded DEAs [32], use the reduction in elastomer thickness and therefore can offer intrinsically linear contractile actuation. For a more systematic understanding of various DEA configurations, one may refer to the review articles [33], [34]. Fig. 3 presents a variety of DEAs and DEA-driven devices, the control of which has been studied.

*B. Electromechanical Behaviors*

The difficulties in the modeling and control of DEAs stem largely from the complex nonlinearities and dynamics inherent in DEAs such as viscoelastic creep, hysteresis and vibrational dynamic effect. Understanding and describing these behaviors is crucial for developing the model-based controllers to meet application requirements, hence these electromechanical behaviors exhibited by DEAs are discussed.

*1) Creep:* DEs as a class of electroactive polymers possess viscoelastic properties due to their unique molecular structure. The mechanical response of a viscoelastic material is time-dependent [35]. The specific manifestations of static viscoelasticity (i.e. a constant stress/strain is applied) include creep and stress relaxation, and under the condition of applying alternating stress the viscoelasticity is manifested as hysteresis and mechanical loss. Creep is a phenomenon that the DEA's output displacement gradually increases with time when subjected to a constant voltage [36]. Fig. 4 shows the creep behavior exhibited by a DEA made of commercial VHB elastomers when a voltage step is applied. As can be seen, there is a slow drift in the output displacement over time after the dynamic region. It has been demonstrated that this phenomenon could be well eliminated by a simple feedforward controller [37] or a conventional PID controller [38].

*2) Hysteresis:* The hysteresis exhibited by DEAs is a non-smooth nonlinear phenomenon between input voltage and output displacement [39], [40], and can be caused by various dissipative processes, such as viscoelastic dissipation, resistance dissipation and polarization loss. The hysteresis nonlinearity has the multivalued character (i.e., the same input voltage can produce different output displacements and the same output displacement can correspond to different input voltages). In addition, it has nonlocal memory effect (the DEA's output displacement depends not only on the current value of input voltage but also on the history of input voltage), resulting in the amplitude dependence of hysteresis [41]. The experiments have demonstrated that the viscoelastic hysteresis in DEAs also exhibits high frequency-dependence [41], [42], [43]. As shown in Fig. 5, not only the area bounded by the hysteresis loop but also the degree of hysteresis asymmetry varies considerably with the excitation frequency. The hysteresis nonlinearity affects the motion accuracy of DEAs, and in severe cases, may even lead to system instability, limiting the practical applications of DEAs [38], [44]. Recent experiments have shown that, for the DEAs made of commercial VHB elastomers, the position error caused by inherent hysteresis can exceed 90% of the travel range [41]. One manifestation of the hysteresis nonlinearity is that the displacement response of a DEA lags behind the input voltage in phase when subjected to cyclic electrical loading [6], [45]. However, it should be pointed out that hysteresis cannot be confused with "phase lag" or "time delay", which can exist in many linear systems [44], [46].

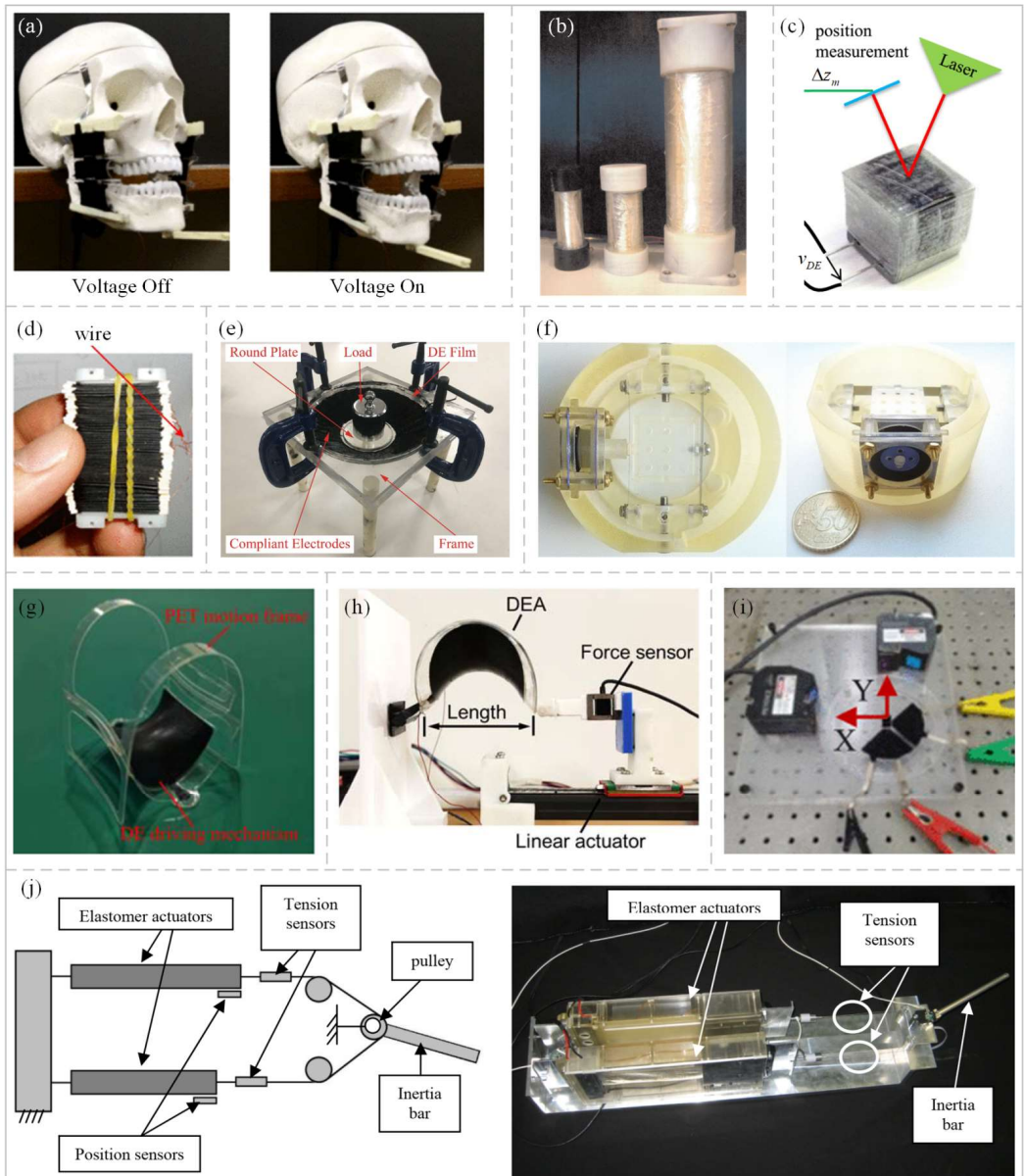

Fig. 3. Various DEAs and DEA-driven devices having been controlled: (a) A robotic jaw driven by two planar DEAs [91]; (b) A rolled DEA [103]; (c) A multilayer stacked DEA [65]; (d) A multi-stacked trapezoid DEA [101]; (e) A conical DEA [76]; (f) A positioning stage driven by two circular out-of-plane DEAs [89]; (g) A soft saddle-shaped DEA [92]; (h) A three-dimensional DEA [115]; (i) A two-degree-of-freedom (2-DOF) DEA [108]; (j) A rotational joint driven by two antagonistic DEAs [100].

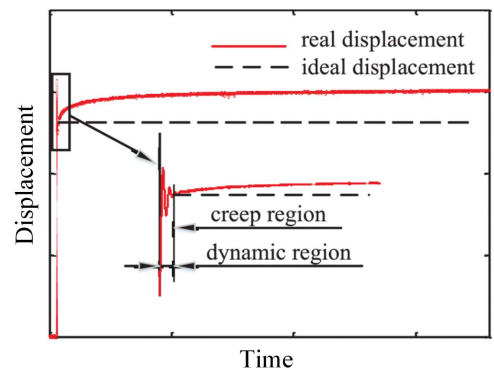

Fig. 4. Step response of a DEA [37].

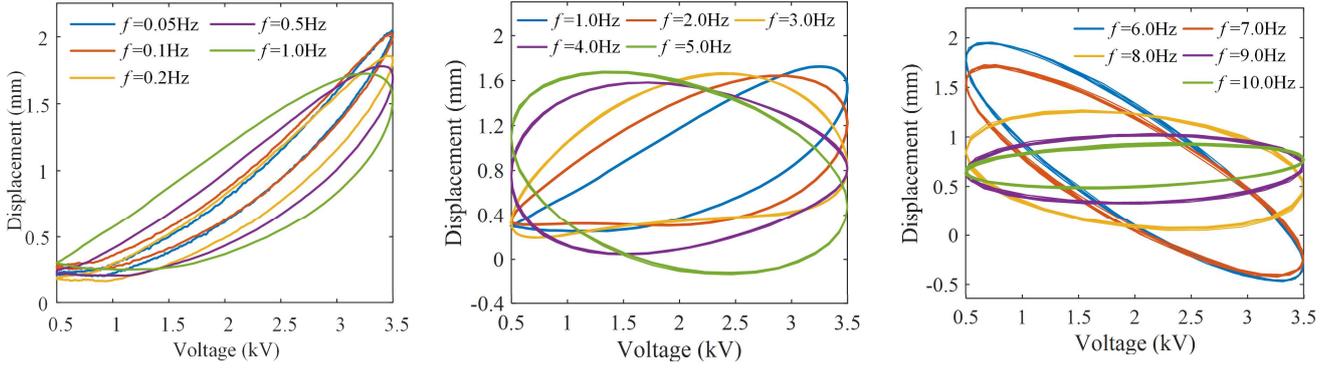

Fig. 5. Hysteresis loops in a DEA under harmonic excitations of different frequencies [41].

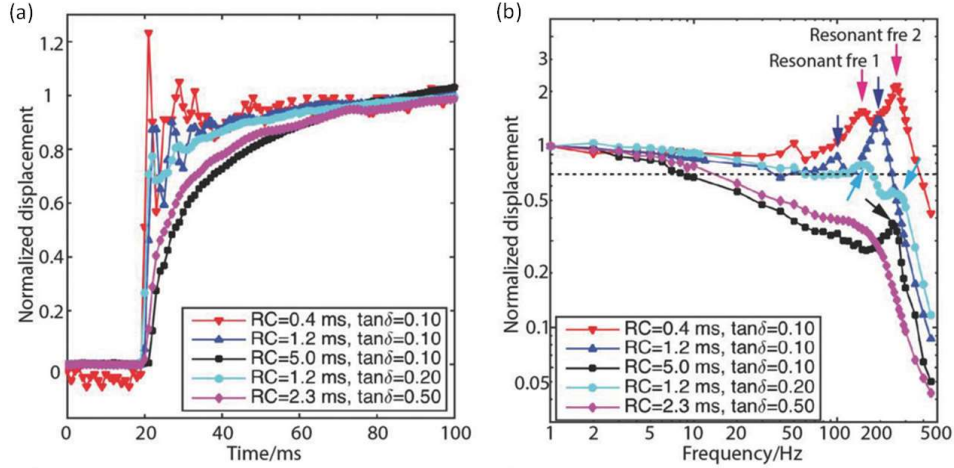

Fig. 6. (a) Step response and (b) frequency response of DEAs based on the combinations of elastomers with varying viscoelasticity (tan δ) and electrodes with varying RC time constants [47].

*3) Vibrational dynamics:* The DEA dynamics means that the effects of the DEA's inertia and damping need to be taken into consideration. Different combinations of dielectric elastomers and electrode materials lead to different values of electrical damping, mechanical damping and stiffness, resulting in DEAs exhibiting distinctly different dynamic responses [47], as shown in Fig. 6. When the electrical losses (including resistance loss and dielectric loss, etc.) and mechanical losses are both low, the DEAs exhibit underdamped responses with overshoots and oscillations for a step input and achieve short rising times (i.e. high response speeds). As with the step response, in the frequency domain, the DEAs with low damping have wide bandwidths and exhibit distinct resonance peaks. It is worth noting that there are two discernable resonant frequencies $f_{r1}$ and $f_{r2}$ in each of these DEAs with a relationship of $f_{r1} = 2f_{r2}$. This is because the induced Maxwell stress is proportional to the square of the applied electric field, leading to exciting the second-order harmonic. Attenuating the higher harmonics caused by nonlinear transduction and amplifying the fundamental harmonic could be achieved by driving the DEA near its natural frequency [4]. When the electrical damping and/or mechanical damping are large, the DEAs exhibit overdamped responses with no overshoots and longer rising times. In terms of frequency response, the DEAs with severe damping have narrow bandwidths and only one or no resonance peak. The stiffness of both the elastomer and the electrodes used in a DEA is a key factor affecting the DEA's natural frequency and therefore its resonant frequency. For instance, the DEAs using a combination of silicone elastomers (such as Elastosil P7670, mixtures of Ecoflex 0030 and Sylgard 184) and carbon nanotube electrodes have resonant frequencies of hundreds of hertz [4], [27], [47], [48], while the resonant frequencies of the DEAs composed of 3M VHB acrylic elastomers and carbon grease electrodes are less than 10 Hz [38], [45]. Developing DEAs with both high bandwidth and large actuation strain to achieve extraordinary power density and energy density exceeding natural muscles has been a research hotspot by exploring new DE materials, electrode materials and fabrication techniques [2], [16], [18].

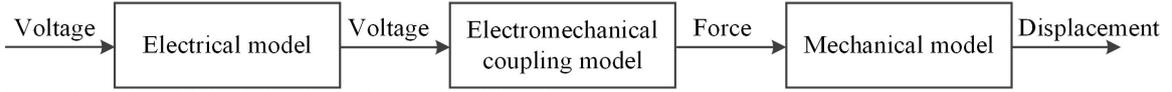

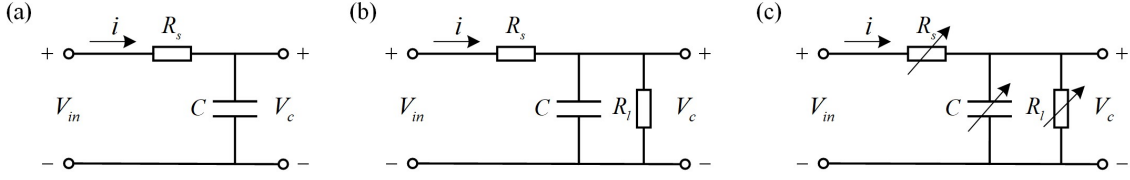

Fig. 7. Different submodels and their relations in physics-based modeling of DEAs.

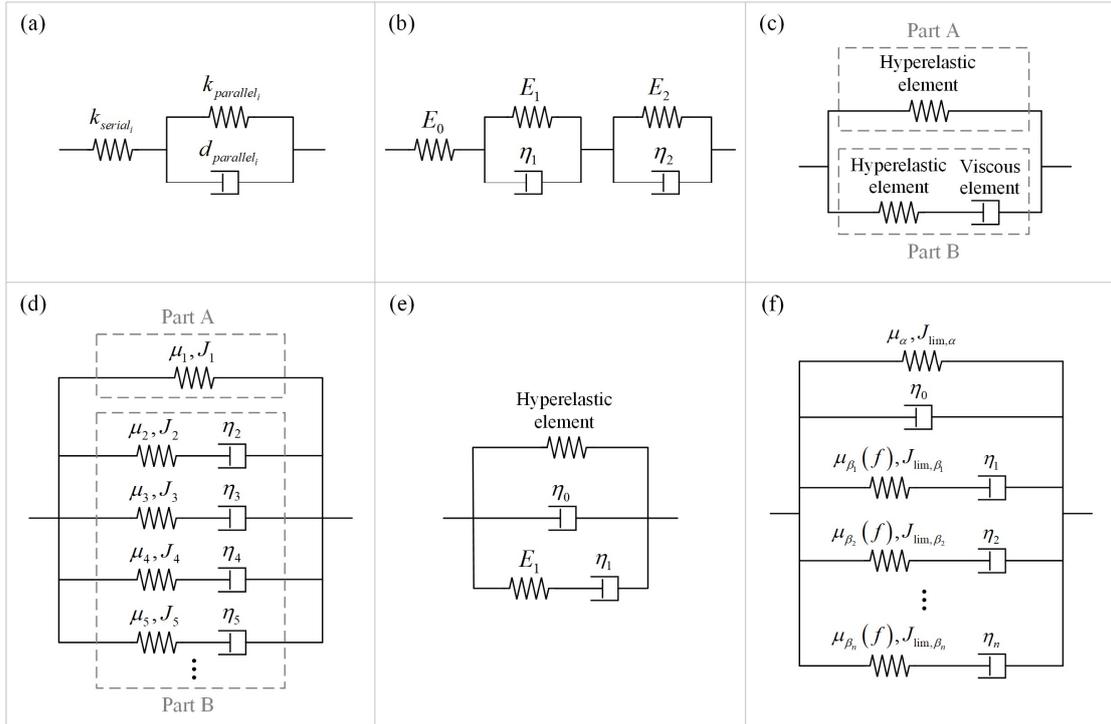

Fig. 8. Equivalent circuit models of a DEA.

Fig. 9. Viscoelastic models for DEs: (a) three-parameter Voigt-Kelvin model [61]; (b) five-parameter Voigt-Kelvin model [36]; (c) a two-part constitutive model [62]; (d) a generalized two-part constitutive model with multiple nonlinear Maxwell elements in parallel [40]; (e) nonlinear Kelvin-Maxwell model [64,65]; (f) generalized nonlinear Kelvin-Maxwell model with frequency-dependent parameters [45].

## III. MODELING METHODS

The aim of electromechanical modeling of soft DEAs is to develop models that can describe the DEA behaviors. The modeling methods for soft DEAs can be roughly divided into two categories: physics-based modeling and phenomenological modeling.

### A. Physics-based modeling

Physics-based electromechanical models of DEAs start from the physical laws and are developed by comprehensively representing different relations involved in DEAs. As shown in Fig. 7, the physics-based model for DEAs generally consists of three submodels: electrical model, electromechanical coupling model and mechanical model. The reviews of the three modeling blocks are as follows.

*1) Electrical models:* In the early stages, a DEA was electrically treated as a parallel plate capacitor [49], [50]. As a matter of fact, the electrodes used in DEAs, including electrode material and electrode thickness, have a non-negligible effect on the actuator performance [27], [47], [51]; on the other hand, dipole polarization causes the dielectric constant of dielectric materials to decrease under high-frequency electric fields. Thus a more reasonable electrical model for the DEA is a series resistor-capacitor circuit, as shown in Fig. 8(a). The series resistance $R_s$ includes the electrode resistance, wire resistance, contact resistance and the equivalent series resistance of polarization loss.

Considering the fact that real dielectric materials are not perfectly insulating and exhibit a certain leakage current, the

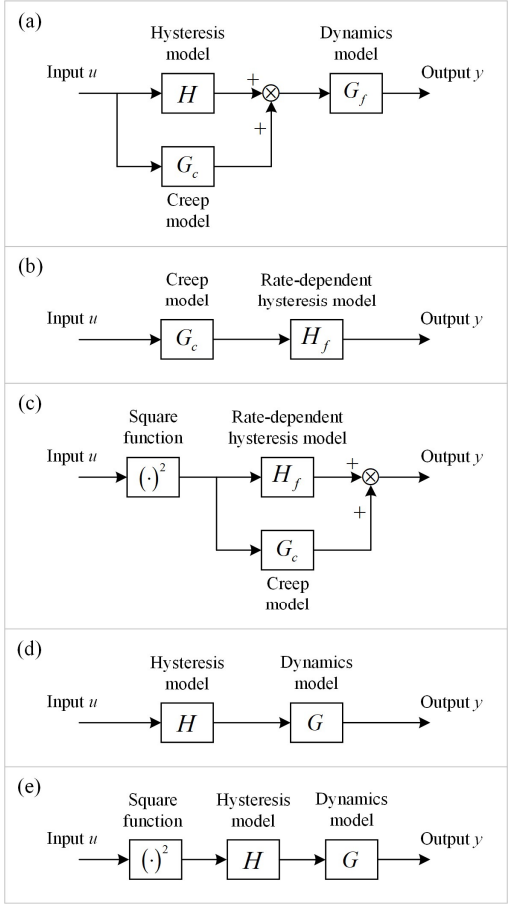

Fig. 10. Block diagram of the model structures for DEAs.

DEA's electrical model includes a parallel leakage resistance $R_l$ in addition to a RC series circuit [36], as shown in Fig. 8(b). Experimental studies have shown the dependence of the capacitance $C$ and the resistances $R_s$ and $R_l$ on the DEA strain [52], [53], [54], [55]. Thus a more realistic circuit model of the DEA is a strain-dependent capacitance connected in parallel with a strain-dependent resistance and in series with a strain-dependent resistance, as shown in Fig. 8(c). If the change in electrode resistance is much smaller than the total series resistance $R_s$, $R_s$ can be assumed to be constant, without introducing a significant modeling error [53]. If only the high-frequency electrical response is of interest, another simplified electrical model is a leakage-free RC series circuit with strain-dependent capacitance and strain-dependent resistance, given that $R_l \gg R_s$ [54], [55].

*2) Mechanical constitutive models:* In the early work, the DE material was assumed to be linearly elastic, which is only valid for small strains [1], [49]. Inspired by the constitutive modeling of rubber materials, hyperelastic models were then used to predict the large-strain behavior of DEs. Kofod adopted three hyperelastic models including Neo-Hookean model, Mooney-Rivlin model and Ogden model to predict the stress-strain behavior of DEs, among which the Ogden model provided a good fit to the experimental data throughout the entire strain range [56]. Wissler and Mazza used Yeoh model, Ogden model and Mooney-Rivlin model to predict the instantaneous response of acrylic DEs under biaxial extension [57]. Gent model can account for the strain-stiffening effect of elastomers by introducing a constant related to the limiting stretch [58], and therefore has also been adopted in many works for describing the mechanical behaviors of DEs [23], [59], [60].

It should be pointed out that linear elastic model and hyperelastic models can only describe the static mechanical behavior. However, DEs as polymeric materials are viscoelastic, that is, possess time-dependent mechanical properties. A common way to model viscoelastic behavior is to combine spring and dashpot elements in a variety of arrangements. Lochmatter et al. [61] used a standard three-parameter Voigt-Kelvin solid model (Fig. 9(a)) to describe the mechanical behavior of DEs. Considering the fact that the three-parameter solid model has one retardation time constant, i.e., can only represent the viscoelastic response of polymers in a short interval of time, Sarban et al. [36] proposed a five-parameter Voigt-Kelvin model for DEs by combining two Kelvin elements and a spring element in series, as depicted in Fig. 9(b). However, the above linear viscoelastic models could only accurately predict the small-strain mechanical behavior of DEs.

Based on the experimental data of stress-strain relation of rubber elastomers under loading conditions of different strain rates, Bergström and Boyce [62] developed a constitutive model (Fig. 9(c)), which consists of two parts connected in parallel: Part A captures the equilibrium response of the elastomer material and can be modeled as any hyperelastic model; Part B captures the time-dependent deviation from the equilibrium state and is modeled as a series connection of a hyperelastic spring and a linear dashpot, which is actually a nonlinear Maxwell element. Foo et al. [63] used the two-part constitutive model to predict the stress-strain relation of DEs under uniaxial tension at different stretching rates, and the results showed good agreement between the model predictions and the experimental data. Then Gu et al. [40] proposed a generalized constitutive model by adopting multiple nonlinear Maxwell elements in parallel, as shown in Fig. 9(d), and the results showed that the model could predict the mechanical response of a DEA subjected to 0.1 Hz cyclic electrical loading. Rizzello et al. [64] and Hoffstadt et al. [65] established a nonlinear Kelvin-Maxwell model (Fig. 9(e)), in which the parallel spring is hyperelastic and is used to capture the quasi-static response. Zhao and Wen [45] proposed a more generalized nonlinear Kelvin-Maxwell model, as shown in Fig. 9(f). The nonlinear Kelvin element with a hyperelastic spring is utilized to describe the steady response of DEs subjected to periodic loading. The other part, i.e. multiple nonlinear Maxwell elements connected in parallel, is used to describe the time-dependent creep behavior. Different from the above viscoelastic models, the modulus of the hyperelastic spring in each nonlinear Maxwell element is assumed to be a linear function of loading frequency. The experimental results showed that this visco-hyperelastic model could predict not only the sweep frequency response of a DEA but also the time-dependent response under sinusoidal electrical loading

TABLE I
SUMMARY OF PERFORMANCE OF DIFFERENT HYSTERESIS MODELS FOR SOFT DEAS

| Model | DE material | Maximum frequency | $e_{rms}$ | $e_m$ | Refs. |
|---|---|---|---|---|---|
| Physics-based model | VHB 4910 | 0.1 Hz | <3% | - | [40] |
| Physics-based model | Silicone (PolyPower) | 10 Hz | - | - | [36] |
| Physics-based model | VHB 4910 | 7 Hz | 3.4% | - | [45] |
| Duhem model | Silicone | 0.1 Hz | 1.05% | - | [80] |
| Butterfly fractional- order Backlash-like model | - | 1 Hz | Min: 0.020 mm Max: 0.036 mm | Min: 2.92% Max: 5.90% | [81] |
| Bouc-Wen model | Silicone | 0.1 Hz | 1.27% | - | [83] |
| Preisach model | VHB 4910 | 0.5 Hz | - | - | [84] |
| Unparallel P-I model | Silicone | 0.7 Hz | Min: 0.88% Max: 3.21% | - | [76] |
| Polynomial-modified P-I model | PDMS | 0.1 Hz | 1.13% | 3.49% | [75] |
| P-I model in cascade with dead-zone operators | Silicone | 0.1 Hz | 1.22% | - | [86] |
| Modified P-I model with rate-dependent elementary operators | VHB 4905 | 1 Hz | Min: 1.82% Max: 4.50% | Min: 4.01% Max: 9.03% | [87] |
| Hammerstein model | VHB 4910 | 10 Hz | Min: 1.92% Max: 5.53% (0.054 mm) | Max: 0.136 mm | [41] |
| Polynomial model | Silicone (PolyPower) | 1.5 Hz | - | - | [42] |
| Ellipse-based model | Silicone (PolyPower) | 1.75 Hz | - | - | [43] |
| LSTM neural network model | PDMS | 0.5 Hz | 1.16% | 2.30% | [88] |

Min: minimum; Max: maximum; $e_{rms}$: root-mean-square error; $e_m$: maximum error.

over a large time scale with good accuracy. In addition to combining spring and dashpot elements to describe the viscoelastic behavior of DEs, Wissler and Mazza [66], [67] employed time-dependent parameters in the hyperelastic strain energy density to account for the time-dependent mechanical response of DEs. The time-dependent hyperelastic parameters $C_{ij}^R$ are defined as $C_{ij}^R(t) = C_{ij}^0 \cdot f(t)$, and the time function $f(t)$ is given as a Prony series, i.e.

$$f(t) = 1 - \sum_{k=1}^{N} g_k \cdot (1 - e^{-t/t_k}) \quad (1)$$

where $C_{ij}^0$ corresponds to the instantaneous elastic response, and $g_k$ and $t_k$ are constants determined from uniaxial relaxation tests of DEs.

*3) Electromechanical models:* A pioneering electro-mechanical modeling work for DEAs was developed by Pelrine et al. [1], [49]. In this work, a DEA is electrically treated as a parallel-plate capacitor, and the electrostatic energy $U$ stored in the DE film can thus be given as

$$U = \frac{Q^2 z}{2\varepsilon A} \quad (2)$$

where $Q$ is the charge on either electrode, $z$ is the thickness of the elastomer film between electrodes, $A$ is the electrode area, $\varepsilon$ is the elastomer permittivity. Applying the elastomer incompressibility constraint (i.e. $Az$ = constant), the effective pressure compressing the DE film in thickness is given as

$$p = \frac{1}{A}\frac{dU}{dz} = \frac{Q^2}{\varepsilon A^2} \quad (3)$$

Assuming the DE is isotropic and its dielectric behavior is linear, the effective pressure $p$ can be further expressed as $p = \varepsilon E^2$, known as Maxwell stress, where $E$ is the electric field. It is noted that $p$ is exactly twice the pressure across a rigid, charged parallel-plate capacitor, which can be explained by the assumption of elastomer incompressibility.

With the assumption of linear elasticity, for small strain cases (the DE film thickness is approximated as the initial thickness), the strain in the thickness direction $s_{z,s}$ can be approximated as

$$s_{z,s} = -\frac{\varepsilon_0 \varepsilon_r (V/z_0)^2}{Y} \quad (4)$$

where $V$ is the applied voltage, $z_0$ is the elastomer's initial thickness, and $Y$ is the elastomer's modulus. For larger strain cases, with maintaining the linear elasticity assumption, the thickness strain $s_{z,l}$ is given by

$$s_{z,l} = -\frac{2}{3} + \frac{1}{3}\left(f(s_{z,s}) + \frac{1}{f(s_{z,s})}\right) \quad (5)$$

where

$$f(s_{z,s}) = \left\{\frac{2 + 27s_{z,s} + \left[-4 + (2 + 27s_{z,s})^2\right]^{1/2}}{2}\right\}^{1/3} \quad (6)$$

The in-plane strains can then be easily determined based on the assumptions of elastomer incompressibility and free boundary. The electromechanical model proposed by Pelrine et al. is simple and but difficult to accurately predict the electrically induced nonlinear large-deformation behavior of DEAs subjected to boundary constraints.

Later, Goulbourne et al. [68] proposed that the total stress in the elastomer σ is the sum of the local elastic stress $\sigma_E$ and the Maxwell stress $\sigma_M$, i.e.

$$\sigma = \sigma_E + \sigma_M \quad (7)$$

where $\sigma_E$ is given as a function of the displacement gradients. Considering the nonlinear large deformation of DEs, the elastomer is mechanically modeled as a hyperelastic solid. Various formulations of hyperelastic model (i.e. strain energy potential) have been developed and substituted into Eq. 7 to predict the voltage-induced deformation of DEAs made of different DE materials. For example, Sarban et al. [11]

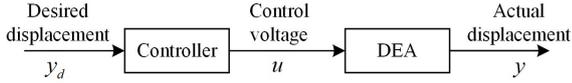

Fig. 11. Block diagram of the open-loop control for DEAs.

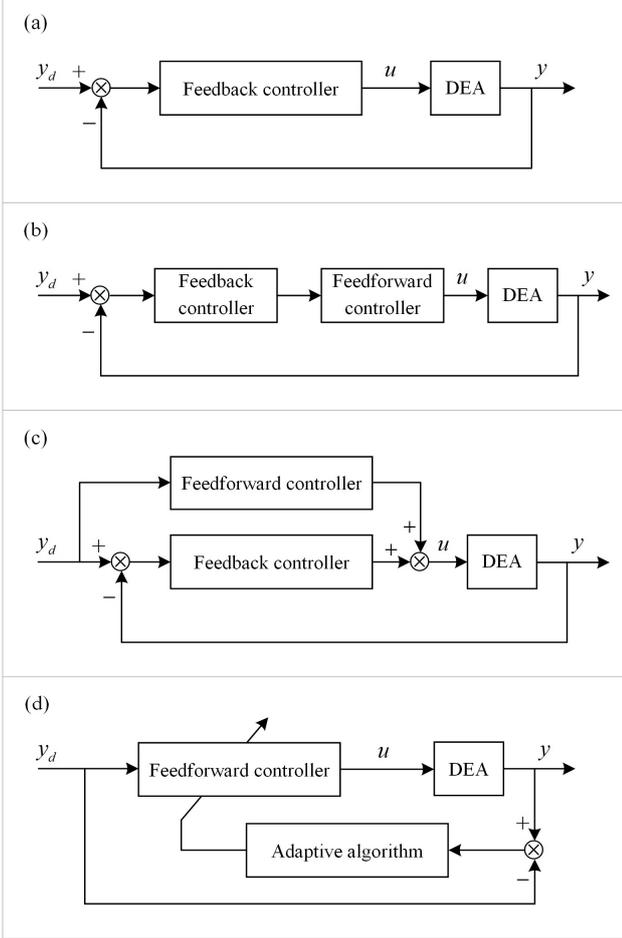

Fig. 12. Different types of control approaches for DEAs ($y_d$ denotes the desired output; $u$ denotes the controller output fed to the DEA; $y$ denotes the actual output of the DEA).

adopted a modified Mooney-Rivlin model and combined it with the stress balance equation to predict the static response of a silicone-based core-free rolled DEA.

Based on the molecular structures of DEs and accumulated research works, Suo and Zhao [69], [70] proposed a general electromechanical modeling framework for DEAs within continuum mechanics and thermodynamics. The basic idea is that when the electrical energy is input, a dielectric elastomer undergoes deformation and its elastic strain energy increases to maintain the stress balance in the elastomer. The energy-based modeling approach has clear physical meaning and its effectiveness has been proven by experiments [40], [71]. Using this modeling approach, for ideal DEs (i.e., incompressibility, isotropy and linear dielectric behavior), the equations of state of the DE are given as follows:

$$\sigma_1 - \sigma_3 = \lambda_1 \frac{\partial W_s(\lambda_1,\lambda_2,\xi_1,\xi_2,\cdots)}{\partial \lambda_1} - \varepsilon E^2 \quad (8)$$

$$\sigma_2 - \sigma_3 = \lambda_2 \frac{\partial W_s(\lambda_1,\lambda_2,\xi_1,\xi_2,\cdots)}{\partial \lambda_2} - \varepsilon E^2 \quad (9)$$

where $\sigma_i$ and $\lambda_i$ are respectively the DE's true stress and stretch ratio along the $i$th direction, and $W_s(\lambda_1,\lambda_2,\xi_1,\xi_2,\cdots)$ represents the elastomer's strain energy density. Once the function $W_s(\lambda_1,\lambda_2,\xi_1,\xi_2,\cdots)$ is given, the electromechanical behavior of a DEA can be analyzed by using Eqs. 8 and 9. For example, when $W_s(\lambda_1,\lambda_2,\xi_1,\xi_2,\cdots)$ was described by various hyperelastic models, the static response of a DEA and the electromechanical instabilities during actuation were comprehensively analyzed by using the above modeling approach [72], [73]. With $W_s(\lambda_1,\lambda_2,\xi_1,\xi_2,\cdots)$ specified by the visco-hyperelastic model shown in Fig. 9(f), the dynamic response of a truncated-cone-shaped DEA was predicted by using the above approach, and the predictions were in good consistency with the experimental results [45].

*B. Phenomenological Modeling*

Different from physics-based models, phenomenological models start from the characteristics of phenomena to produce the system behaviors, without necessarily providing physical insights into the phenomena. Various models have been developed to describe the DEA behaviors and can be broadly divided into five categories according to the model structure with block connection, as shown in Fig. 10. The first model structure is that hysteresis, creep and vibrational dynamics are each modeled as a separated block, as shown in Fig. 10(a). The second and third model structures involve directly establishing rate-dependent hysteresis models, that is, considering the hysteresis nonlinearity and rate-dependent effects of DEAs together. In the second model structure, a creep model is cascaded with a rate-dependent hysteresis model, as depicted in Fig. 10(b). In the third model structure (shown in Fig. 10(c)), a creep model and a rate-dependent hysteresis model are connected in parallel, and a square function is connected in series in front of them, considering that the DEA response directly depends on the square of driving voltage. In the fourth model structure, the creep and vibration are modeled together as a dynamic model, which is connected in series with a hysteresis model, as shown in Fig. 10(d). Compared with the fourth model structure, a square function module is added in the fifth model structure so as to take into account the quadratic input characteristic of DEAs, as shown in Fig. 10(e). A detailed discussion of the different modeling blocks is given below.

*1) Creep models:* In order to describe creep (a slow drift phenomenon), linear models and nonlinear models (also called logarithmic models) have been proposed. It should be pointed out that linear models are generally used to describe the creep in DEAs. Zou et al. [37] defined the relative creep displacement $C_r(t)$ as the ratio of the displacement creep amount to the ideal displacement in the case of no creep. The ideal displacement $y_0(t)$ is assumed to have the following relationship with the step input voltage $v$: $y_0(t) = Kv$, where $K$ is the proportionality coefficient dependent on the input voltage. $C_r$ is modeled as a three-order linear transfer function, with the poles located at $-\lambda_i < 0$, where $y_{c,i}(t)$ denotes the output of the $i$th creep operator. The discrete-time form of $y_{c,i}$ with sampling period $T$ is thus expressed as

$$y_{c,i}[v](k) = e^{-\lambda_i T} y_{c,i}[v](k-1) + (1 - e^{-\lambda_i T})v(k-1) \quad (12)$$

and the creep model is then represented as

$$y_c[v](k) = \sum_{i=1}^{n_c} c_i y_{c,i}[v](k) \quad (13)$$

where $c_i$ are weight coefficients and $n_c$ is the number of creep operators. It should be pointed out that this creep model is essentially a generalized Kelvin model.

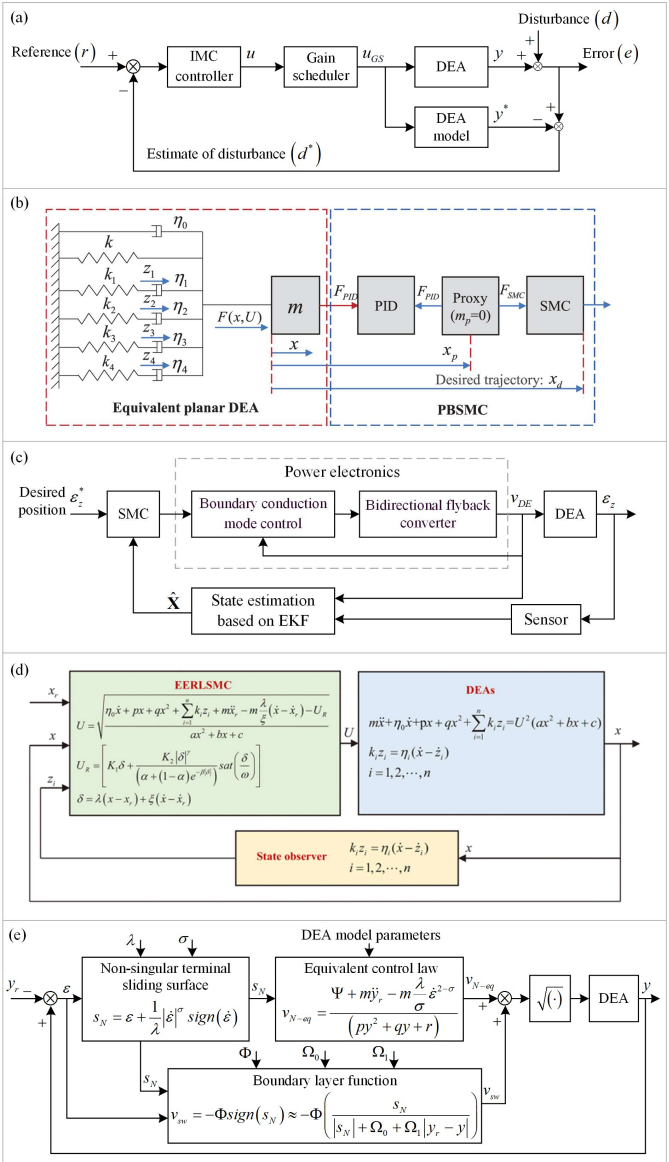

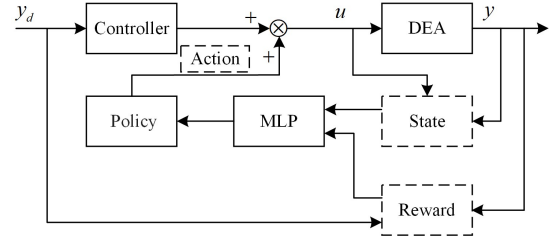

Fig. 14. Schematic of the control method based on deep reinforcement learning.

*2) Hysteresis models:* Developing an accurate hysteresis model is critical for a system to achieve hysteresis compensation and precise control. The complex characteristics of hysteresis increase the difficulty of hysteresis modeling. A number of hysteresis models have been developed in the literature and can be roughly classified into physics-based models and phenomenological models. The representative examples of physics-based models are Jiles-Atherton model for describing ferromagnetic hysteresis [77] and a domain wall model for describing piezoelectric hysteresis [78]. For DEAs, some physics-based models have been proposed (see Section III-A for details) and experimentally proven to describe the electromechanical hysteretic behavior [36], [40], [45]. These physics-based models are generally complex, making further controller design challenging. For example, 14 parameters and 21 parameters are needed to be identified respectively for the models in [40] and [45] when four nonlinear Maxwell units are used to ensure modeling accuracy. In addition, a physics-based model developed for a DEA of one DE material or one configuration may not be applicable to another.

Different from physics-based models, the phenomenological models for describing hysteresis cannot provide physical insights into the hysteresis phenomenon. According to the mathematical structure, the phenomenological hysteresis models can be further categorized into differential-equation based hysteresis models, operator-based hysteresis models and others.

Differential-equation based hysteresis models include Duhem model, Backlash-like model and Bouc-Wen model. The Duhem model, firstly developed in [79], was used in [80] to describe the asymmetric hysteresis in DEAs. In order to describe the butterfly-shaped hysteresis loops exhibited by DEAs subjected to unbiased sinusoidal voltages, Li et al. [81] developed a butterfly fractional-order Backlash-like hysteresis model by introducing a twist mechanism. In this model, a

Fig. 13. Model-based feedback control schemes for control of soft DEAs. Adapted from [98], [106], [65], [108], [109].

and according to the definition of $C_r$, the creep model can be given by

$$y(t) = Kv(1 + C_r(t)) \quad (10)$$

Based on the relative creep model and the assumption that the relative creep displacement and relative creep voltage are in a linear relationship, a creep compensator was designed to eliminate the creep in DEAs [37], [74].

In addition, creep can also be modeled as a linear superposition of several fundamental creep operators [75], [76]. Each creep operator is defined as a first-order linear system:

$$\frac{1}{\lambda_i} \dot{y}_{c,i}(t) + y_{c,i}(t) = v(t) \quad (11)$$

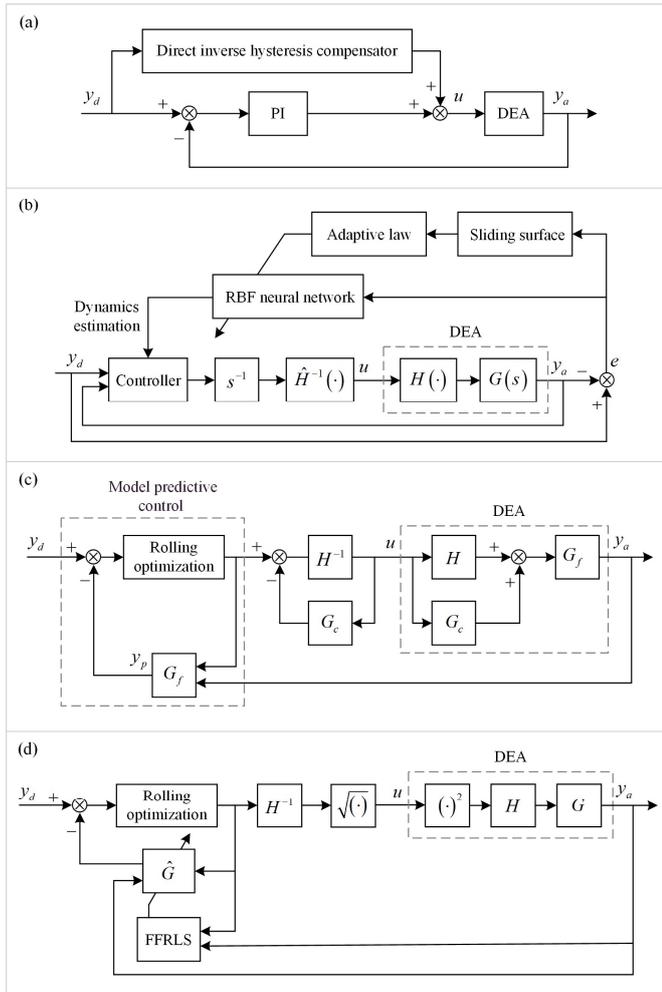

Fig. 15. The first type of feedforward-feedback control methods for soft DEAs. FFRLS: forgetting factor recursive least squares.

fractional-order differential calculus was introduced to improve the model accuracy, and given the insensitive zone in DEAs, a threshold selector was applied. The Bouc-Wen model, proposed initially by Bouc and further generalized by Wen [82], was also utilized in [83] to describe the hysteresis of DEAs.

Operator-based hysteresis models are the models with an integral of weighted elementary hysteresis operators. Common operator-based hysteresis models for smart materials based actuators include Preisach model, Prandtl–Ishlinskii (P-I) model and Krasnosel'skii–Pokrovkii model. Among them, so far only the Preisach model with Relay operators and the P-I model with Play operators have been used to describe the hysteresis in DEAs.

The Preisach model, first developed for hysteresis modeling in ferromagnetic materials [39], has a general mathematical structure and therefore has been utilized for describing the hysteresis loops in DEAs [84]. Since the analytical inverse solution of the Preisach model is difficult to obtain, numerical methods are usually required to obtain an approximate model inverse for hysteresis compensation [84].

The classical P-I model consists of a linear function of the input and weighted Play or Stop operators. Compared with the Preisach model, the P-I model has a simpler structure and is analytically invertible, making it more commonly used in the modeling and control of DEAs. However, the Play operator and Stop operator have symmetric and rate-independent properties, resulting in the classical P-I model being unable to describe the asymmetric hysteresis loops in DEAs. In order to overcome this limitation, some modified P-I models have been developed for asymmetric hysteresis description. Zhang et al. [76] employed the extended unparallel P-I model with unparallel Play operators developed in [85] to model the asymmetric hysteretic phenomena in DEAs. Compared with the classical Play operator, the left descending edge of the unparallel Play operator is able to tilt freely by multiplying a factor. Without modifying the classical Play operator, Huang et al. [75] established a generalized P-I model by using a high-order polynomial function of the input to replace the linear function of the input. Wang et al. [86] adopted the idea of cascading the dead-zone operators with the Play operators to establish an asymmetric P-I model. In the above modified P-I models, the one-side Play operator was adopted instead of the two-side Play operator, considering that the output displacement of a DEA is positive.

It is worth noting that classical operator-based hysteresis models and the above modified P-I models are amplitude-dependent but rate-independent. However, the hysteresis in DEAs exhibits a high degree of rate-dependence, as depicted in Fig. 5. To overcome the limitation that the operator-based hysteresis models cannot describe the rate-dependent hysteresis, some modifications need to be made. From a mathematical perspective, there are two feasible approaches: redefining an elementary hysteresis operator and developing a dynamic density function. Zou et al. [87] proposed a modified rate-dependent P-I hysteresis model for DEAs, in which a fourth-order polynomial function of the input was utilized to capture the asymmetric property and a rate-dependent elementary operator was developed based on two dynamic envelope functions incorporating the input voltage and its derivative with respect to time. This model well reproduced the hysteresis loops of a DEA made of VHB at sinusoidal voltages within the range of 1 Hz. However, it should be noted that due to the introduction of the input's time derivative, the elementary hysteresis operator becomes more complicated, thereby increasing the difficulties of model identification and hysteresis compensator design. Unlike the above rate-dependent hysteresis model constructed by directly modifying the elementary hysteresis operator, Zhao et al. [41] employed the Hammerstein modeling approach with a cascaded structure to describe the highly rate-dependent and asymmetric hysteresis behavior of DEAs. The static nonlinear element was represented by a modified P-I model incorporating a three-order polynomial function of the input voltage to describe the asymmetric hysteresis nonlinearity, while the dynamic element was represented by an adaptive filter to capture the frequency dependence of hysteresis. It has been shown that the model could predict the hysteresis loops of a DEA at voltages

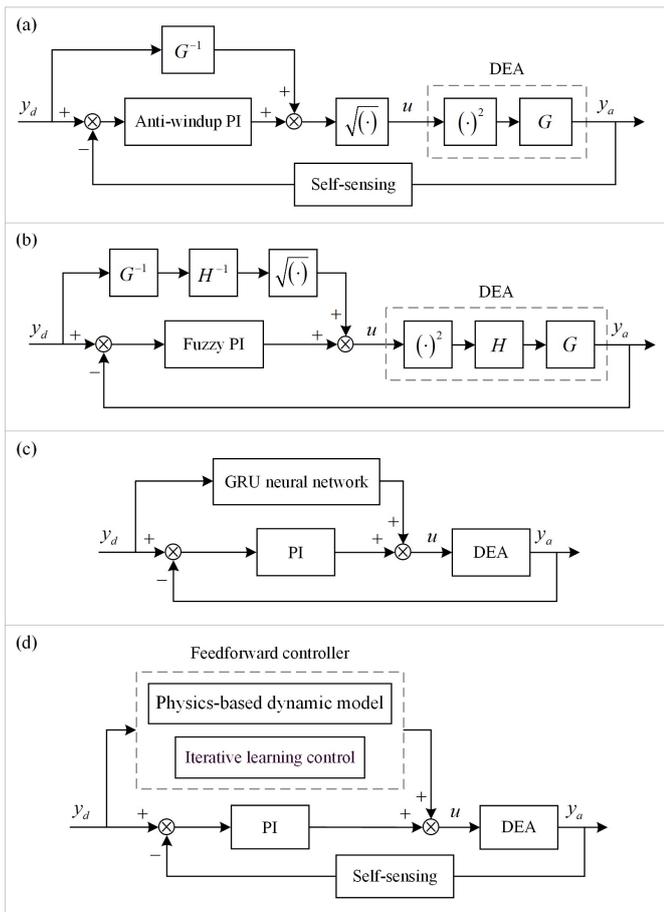

Fig. 16. The second type of feedforward-feedback control methods for soft DEAs.

of different frequencies and different amplitudes with good accuracy.

Apart from differential-equation based hysteresis models and operator-based hysteresis models, there are also other methods for modeling the hysteresis of DEAs. Zhang et al. [42] used two polynomials to model the rising and falling branches of hysteresis loops. The advantage of this modeling approach is that the mathematical formula is simple, but the orders and coefficients of the two polynomial functions all need to be re-determined for the hysteresis loops at different frequencies, amplitudes and waveforms. Besides, Tian et al. [43] developed an ellipse-based hysteresis modeling approach to describe the hysteresis of DEAs. Given that the asymmetric characteristic of hysteresis loops is because the actuator displacement is related to the square of the input voltage, a linearization process was firstly performed to make the shape of hysteresis loops symmetric. The resulting symmetric hysteresis loops were then modeled as elliptical equation models in the constrained conic form and the general parametric form. In addition, neural networks consisting of interconnected artificial neurons with adjustable weights, have demonstrated potential in modeling nonlinear systems. Considering the memory characteristic of hysteresis behavior, Xiao et al. [88] employed the long-short term memory (LSTM) neural network after training to model the hysteresis of DEAs. The comparison of different hysteresis models developed for DEAs is given in Table I. It is worth mentioning that although many hysteresis models for DEAs have been developed, there is currently no general hysteresis model that can precisely describe the hysteresis nonlinearity of DEAs made of various materials.

*3) Dynamic models:* Dynamic models are used to describe the vibration, overshoot and frequency-dependence behaviors of a system. The dynamics of a DEA is generally modeled as a linear system. The order of the linear dynamic model is determined by comprehensively considering the modeling accuracy and the computational complexity of model parameter identification. The model parameters can be identified by employing different approaches such as ordinary least-squares approach [89], nonlinear least-squares approach [75], [80], [86] and Hopfield neural network approach [83].

## IV. CONTROL METHODS

The control of DEAs aims to meet the motion or positioning requirements of DEAs in different applications. However, the precise control of soft DEAs is difficult due to the inherent nonlinearities of DEAs including viscoelastic creep, hysteresis and quadratic-input nonlinearity as well as complex dynamics. Especially, the hysteresis, as a non-smooth nonlinearity, will reduce the motion accuracy and the system instability, making the controller design quite challenging. Various control methods for DEAs have been developed in the past few years and can be roughly classified into two categories: open-loop control and closed-loop control according to whether the feedback loop is used. In the following, these two types of control methods for DEAs will be detailed and discussed.

### A. Open-loop Control

The current open-loop control for DEAs is simple feedforward control, which utilizes the known system models to obtain the control input. When a DEA is modeled as a physics-based model, a feedforward inverse controller could be designed by deriving the function of input voltage with respect to desired displacement [90], [91], [92]. When a DEA is modeled as various phenomenological models with cascaded structures as shown in Fig. 10, the basic idea of feedforward control is to construct corresponding compensators for the modeling blocks and place them in the feedforward path so as to mitigate the undesirable behaviors, as depicted in Fig. 11.

For creep compensation, with using the creep model expressed as Eq. 10, a feedforward compensator was developed based on the idea that the relative creep voltage and the relative creep displacement have the same creep characteristics but different creep factors [37], [74]. Using the linear creep model expressed as Eq. 13, a feedforward inverse compensator was constructed by obtaining the analytical inverse of the established creep model and multiplying it by a second-order inertial link to ensure physical realizability [76]. It should be noted that the feedforward compensation for creep generally no longer needs to be implemented in the closed-loop control, as traditional feedback controllers such as a simple PID controller can easily mitigate the creep [38].

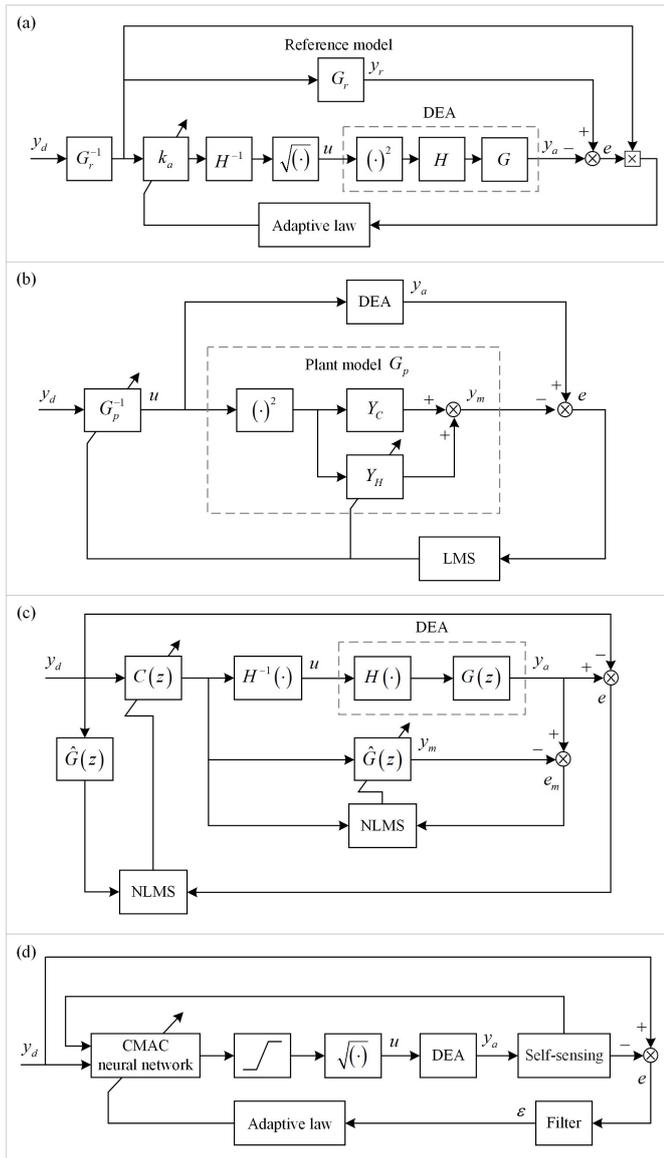

Fig. 17. Different adaptive control schemes for DEAs.

Various hysteresis models (detailed in Section on III-B-2) have been developed to describe the hysteresis nonlinearity of DEAs. There are generally three methods to design a feedforward compensator based on the established hysteresis model. The first method is to obtain an inverse of the employed hysteresis model as the feedforward compensator. Qiu et al. [93] used the analytical inverse of classical P-I model, firstly derived by Krejci and Kuhnen [94], and placed it in the feedforward path to compensate for the hysteresis of DEAs. When the hysteresis was modeled as a cascade of Play operators and dead-zone operators, the analytical inverse of this hysteresis model could be obtained by utilizing the analytical inverse solutions of the classical P-I model and the dead-zone model [86]. Zhang et al. [84] proposed a searching algorithm to obtain an approximate inverse solution of the Preisach model for hysteresis compensation. In addition, the inverse of the Duhem model was derived as a feedforward compensator to eliminate the hysteresis behavior of DEAs [80]. The second method is to directly utilize the established hysteresis model for describing the inverse hysteresis, without requiring model inversion. This method is based on the idea that the inverse hysteresis is essentially a type of hysteresis loop from a mathematical point of view. The difference between the inverse hysteresis and the actual hysteresis in a DEA lies in the orientation of hysteresis loops. With this method, Zou et al. developed two direct inverse hysteresis compensators using a modified rate-independent P-I model and a modified rate-dependent P-I model respectively [38], [74]. Moreover, the polynomial model [42] and the gated recurrent unit (GRU) neural network [95] were also directly used to model the inverse hysteresis of DEAs. It should be noted that the parameters of the inverse hysteresis model are different from those of the hysteresis model, and thus need to be re-identified. The third method is to utilize the established hysteresis model to construct a hysteresis compensator, without finding the inverse of the hysteresis model. Based on the Bouc-Wen hysteresis model, Rakotondrabe [96] designed a hysteresis compensator with an inverse multiplicative structure by only using the linear term to extract the control signal. Later, this hysteresis compensator was used to compensate for the hysteresis nonlinearity in DEAs [83]. Compared with the first and second methods, this method has the following advantages: (1) no hysteresis model inversion is required; (2) no additional computation or identification work is required. Using the similar idea to that in [96], Huang et al. [75] and Zhang et al. [76] respectively designed feedforward hysteresis compensators by dividing the modified P-I models into the classical P-I model and others first and then inverting the classical P-I model to extract the control signal.

The quadratic nonlinearity associated with the electromechanical transduction principle of DEAs can be eliminated by inserting a square root function [64], [76], [80], [83], [86], [89], [97], [98]. Developing a feedforward controller for compensating the vibration of DEAs is in essence inverting the system dynamics model [80], [86], [89]. As mentioned in Section III-B-3, the dynamic behavior of a DEA system is generally described by a linear dynamic model; when the model is represented by a transfer function, the model inverse can be directly obtained. However, it is worth noting that the model inverse is physically unrealizable because its denominator order is lower than its numerator order. This requires that the model inverse be multiplied by an inertia link of sufficient order so as to develop a vibration compensator with a proper transfer function [80], [86]. In addition, the input-shaping method is also used for the vibration reduction of DEAs. For example, Zou et al. [37] employed the zero vibration input-shaping technique to design a vibration compensator, which successfully suppressed the oscillation and overshoot of a DEA under step excitation.

It is important to note that the effectiveness of a feedforward controller depends on the accuracy of the developed system model. Additionally, the open-loop control is unable to correct the positioning or motion errors resulting from model uncertainties and external disturbances. Therefore,

TABLE II
SUMMARY OF PERFORMANCE OF DIFFERENT CONTROL METHODS FOR SOFT DEAS

| Control method | | DE material | Maximum trajectory frequency | Maximum travel range | Robustness | Refs. |
|---|---|---|---|---|---|---|
| Type | Control technique | | | | | |
| Feedforward | Inverse | Natural rubber, Oppo Band | 5 Hz | 0.11 mm | - | [90] |
| | Inverse | VHB 4905 | 0.5 Hz | 11 mm | - | [91] |
| | Inverse | VHB 4910 | - | 10 mm | - | [92] |
| | Direct compensation | VHB 4905 | 1.5 Hz | 3 mm | - | [74] |
| | Inverse | Silicone | 60 Hz | 80 μm | - | [89] |
| Feedback | PID | Silicone (PolyPower) | 0.25 Hz | 0.6 mm | - | [100] |
| | Digital PID with an integrator anti-windup | - | - | 1.6 mm | - | [102] |
| | Pulsewidth-modulated PID | Synthetic rubber | 0.5 Hz | 0.3 mm | - | [101] |
| | neurofuzzy PD+I | PVDF | 4 Hz | 3 mm | ✓ | [104] |
| | Square root function + $H\infty$ robust PID | Silicone | - | 0.5~0.6 mm | ✓ | [97] |
| | Square root function + robust PID based on linear matrix inequality optimization | Silicone | 0.1 Hz | 2~3 mm | - | [64] |
| | Square root function + internal model control | Silicone (PolyPower) | - | 1.5%~2% | ✓ | [98] |
| | Proxy-based SMC | Wacker Elastosil 2030 | 4 Hz | 0.2 mm | - | [106] |
| | SMC based on enhanced exponential reaching law | Wacker Elastosil 2030 | 5 Hz | 0.3 mm | - | [108] |
| | | VHB 4910 | 4 Hz | 4 mm | | |
| | Nonsingular terminal SMC | VHB 4910 | 5 Hz | 1.5 mm | - | [109] |
| | Deep Q-learning with MLP | VHB 4910 | ≈0.1 Hz | 4 mm | ✓ | [111] |
| Feedforward-feedback | Direct hysteresis compensation + PI | VHB 4905 | 1 Hz | 3 mm | - | [38] |
| | Hysteresis inverse + SMC + RBF neural network | - | 1/(2π) Hz | 2 mm | ✓ | [93]* |
| | Hysteresis pseudoinverse + dynamic surface control | VHB 4910 | 0.15 Hz | 0.6 mm | ✓ | [84] |
| | Inverse compensation of hysteresis and creep + model predictive control | PDMS | 0.25 Hz | 1 mm | - | [75] |
| | Inverse compensation of quadratic nonlinearity and hysteresis + generalized predictive control | Silicone | - | 0.5 mm | - | [83] |
| | Inverse compensation of quadratic nonlinearity and linear dynamics + anti-windup PI | Wacker Elastosil 2030 | - | 0.5~1 mm | - | [112] |
| | Inverse compensation of quadratic nonlinearity, hysteresis and linear dynamics + fuzzy PI | Silicone | - | 1 mm | - | [86] |
| | Direct compensation based on GRU neural network + PI | Silicone | 0.4 Hz | 1 mm | - | [95] |
| | Iterative learning control + PI | Silicone | 1.4 Hz | 1~1.5 mm | - | [113] |
| Adaptive feedforward | Inverse compensation of quadratic nonlinearity and hysteresis + MRAC | Silicone | - | 1 mm | - | [80] |
| | Inverse compensation of quadratic nonlinearity, hysteresis and creep + LMS | Silicone | - | 1 mm | ✓ | [76] |
| | Direct hysteresis compensation + filtered-x NLMS | VHB 4910 | 10 Hz | 1.4 mm | - | [114] |
| | Square root function + CMAC neural network | VHB 4910 | 0.2 Hz | 10 mm | ✓ | [115] |

*Simulation

for the applications where the high-precision control of DEAs is required, the closed-loop control with feedback loop is generally used.

*B. Closed-loop Control*

Compared with open-loop control, the closed-loop control with feedback can utilize the real-time output information to adjust the control signal accordingly, thereby ensuring the control performance of systems in the presence of unknown disturbances and model uncertainties. On the other hand, owing to the existence of feedback, the closed-loop control has a certain degree of reduction in the requirement of modeling accuracy. However, the stability of the closed-loop control system needs to be considered. The closed-loop control approaches for DEAs can be broadly divided into three categories: feedback control, feedforward-feedback control and adaptive feedforward control, as illustrated in Fig. 12. The following is a detailed introduction and discussion of these control approaches.

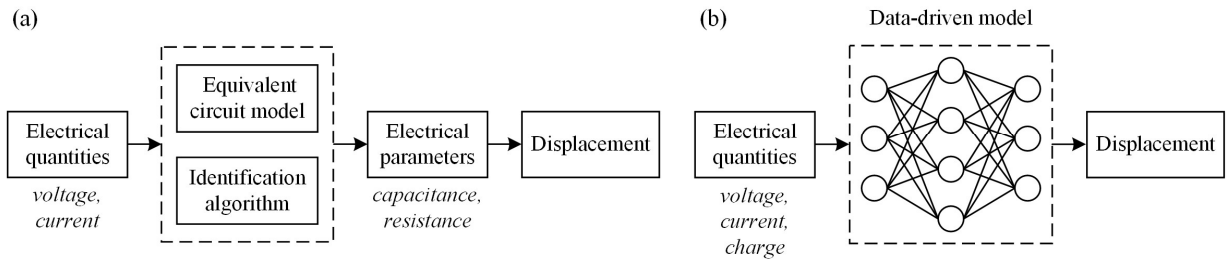

Fig. 18. Self-sensing approaches for DEAs.

*1) Feedback control:* Fig. 12(a) shows the block diagram of the feedback control for DEAs. The early attempts have been concentrated on PID control, which is the simplest and most commonly used control method in industrial applications. Various PID controllers, including classical PID controller [99], [100], pulsewidth-modulated PID controller [101] and digital PID controller with an integrator anti-windup scheme [102], have been applied for the control of DEAs. In order to reduce the high-frequency effects introduced by the derivative action and to eliminate the quadratic nonlinearity in the driving voltage, a linear first-order filter was cascaded with a PID controller and a square root function was inserted between the controller and the DEA [97]. Taking into account the model uncertainties, this approach was combined with the standard direct loop-shaping $H_\infty$ robust control design technique to obtain a better tracking performance. It should be pointed out that the PID controller parameters were determined based on the inverse of the linear model of a DEA, which was obtained by developing a nonlinear dynamic model incorporating viscoelasticity and electromechanical coupling effect and then linearizing it around a predefined equilibrium point corresponding to a constant voltage input. Foreseeably, the controller based on model linearization had satisfactory performance only around the linearization point. To address this limitation, a physics-based nonlinear dynamic model for DEAs was firstly established with the same modeling idea in [97], and then reformulated as a linear parameter-varying system. By means of linear matrix inequality optimization algorithms, a robust controller having the form of a PID control law was obtained, which guaranteed worst case performance in the whole operating range [64].

In addition to PID control, internal model control (IMC) scheme was also used for DEA control. Jones and Sarban [98] designed an internal model controller by using the inverse of a DEA model and a gain scheduling term in series (Fig. 13(a)). Using the same IMC approach in [98] and plant model reduction, a model-based PID controller could be designed according to the transfer function of the equivalent closed-loop feedback controller [103]. There have also been efforts to use model-free intelligent control techniques such as fuzzy logic PD+I control and neurofuzzy PD+I control for the control of DEAs [104]. It should be noted that these control schemes are generally limited to position control and ultra-low frequency trajectory tracking of DEAs because of neglecting the hysteresis nonlinearity.

To further address the limitations, some attempts have been made to design control laws with modern control techniques. Among them, sliding mode control (SMC) is a popular nonlinear robust control technique for DEAs owing to its robustness to parameter variations and disturbances as well as its ease of implementation. A proxy-based SMC scheme, firstly developed in [105], was employed for the tracking control of DEAs [106]. As depicted in Fig. 13(b), the actual controlled object was connected to a virtual object (referred to as a proxy) through a PID controller, which ensured the position of the end-effector to follow the position of the proxy. A sliding mode controller, acting on the other side of the proxy, ensured the proxy to track the desired trajectory. The key feature of this proxy-based SMC is the introduction of a virtual proxy, which allows the responses to large tracking error and small tracking error to be designed independently. Based on the established holistic electromechanical model, Hoffstadt and Maas [65] used SMC for position regulation of DEAs, in which the state variables were estimated using an extended Kalman filter (EKF) (Fig. 13(c)). It should be pointed out that the conventional SMC suffers from the chattering phenomenon near the sliding surface due to the discontinuous function in SMC. In order to mitigate chattering effects and reduce the time required for the system trajectory to reach the equilibrium point, a SMC scheme based on enhanced exponential reaching law was developed in [107] and then applied to DEAs [108], as shown in Fig. 13(d). To achieve a better balance of tracking accuracy and chattering suppression and to guarantee finite-time convergence, a nonsingular terminal SMC combined with an enhanced boundary layer switching function was proposed for DEAs [109], as shown in Fig. 13(e). To cope with the system uncertainty, the adaptive SMC approach combining SMC and adaptive control has been developed for DEAs, where the adaptive control was introduced to estimate the system parameters and the adaptive law was designed by the Lyapunov method to ensure the system stability [110].

In addition, there have been attempts to use reinforcement learning techniques for the control of DEAs, considering the fact that accurately modeling DEAs is difficult, especially for those with complex configurations. Li et al. [111] proposed a model-free control method based on deep Q-learning to achieve accurate time-series control of DEAs, in which the Q-

TABLE III
COMPARISON OF DIFFERENT DISPLACEMENT SELF-SENSING METHODS FOR DEAS

| Driving voltage | Electrical quantity | Capacitance estimation method | Displacement estimation method | Accuracy | Refs. |
|---|---|---|---|---|---|
| A LF, HA actuation voltage + a HF, LA sensing voltage | Voltage, current | Impedance analysis | Physics-based model | ≤1.93% | [55] |
| A HV DC actuation voltage + a small time-varying sensing voltage | Voltage, current | Multidimensional linear regression algorithm | Physics-based model | - | [121] |
| A LF, HA actuation voltage + a HF, LA sensing voltage | Voltage, current | Impedance analysis | Piecewise linear fitting | 0.05 mm | [118] |
| A LF, HA actuation voltage + two HF, LA sensing voltages | Voltage | Impedance analysis | Autoregressive exogenous model | ≤0.7 mm | [115] |
| A LF, HA actuation voltage + a HF, LA sensing voltage | Voltage, current | FFRLS algorithm | Polynomial fitting | ≤1.89% | [116] |
| A LF, HA actuation voltage + a HF, LA sensing voltage | Voltage, current | FFRLS algorithm | Polynomial surface fitting | 0.069 mm | [112] |
| Any arbitrary driving voltage | Voltage, current | EKF | Physics-based model | <4.5% | [120] |
| A LF, HA actuation voltage + a HF, LA sensing voltage | Current | - | Polynomial fitting | 6.43% | [122] |
| | | | BP neural network | 2.74% | |
| | | | BP neural network with data pre-processing | 2.32% | |
| | | | NARX neural network | <1% | |
| - | Voltage, current, charge | - | NARX neural network | <0.15% | [113] |

LF: low frequency; HF: high frequency; LA: low amplitude; HA: high amplitude; HV: high voltage; DC: direct current.

function is parametrized by a multilayer perceptron (MLP), as shown in Fig. 14. The data-driven control scheme not only does not require any knowledge of the structure or material parameters of DEAs, but also can learn to track new trajectories by learning control policies from training data. The robustness of this control method to the changes in structure and material property of DEAs has been verified experimentally.

*2) Feedforward-feedback control:* The feedforward-feedback control integrates the merits of feedforward control and feedback control and is therefore widely used for high-precision trajectory tracking of DEAs. The feedforward control and feedback control can be combined in series or in parallel, as depicted in Fig. 12(b) and (c). Depending on what the feedforward controllers compensate for, the feedforward-feedback control approaches for DEAs can be roughly divided into two types: (1) the feedforward controllers compensate for hysteresis nonlinearity but not vibrational dynamics; (2) the feedforward controllers compensate for both hysteresis and vibrational dynamics.

In the first type of feedforward-feedback control approach, the feedforward controllers are designed for compensating the hysteresis and even creep and quadratic input nonlinearity but not including dynamics, and various feedback controllers are then developed for the compensated dynamic system with uncertainties. Zou et al. [38] designed a direct inverse hysteresis compensator using a modified rate-independent P-I model and adopted a conventional proportional-integral feedback controller to deal with the creep effect and model uncertainties (Fig. 15(a)). It should be pointed out that this work, though pioneering, does not take into account the system dynamics. With the DEA modeled as a cascade of a hysteresis model and a second-order linear dynamic model, a hysteresis compensator based on the analytical inverse of P-I model was adopted, and a sliding mode controller combined with radial basis function (RBF) neural network was designed to reduce the effects of residual hysteresis and system uncertainties [93], as shown in Fig. 15(b). In this control scheme, the learning law of the RBF neural network weights was determined by the Lyapunov method so as to ensure the asymptotic stability of the closed-loop system. However, this control method has only shown good tracking control results in simulations and has not been verified experimentally. Another attempt is a dynamic surface controller based on barrier Lyapunov function in combination with a hysteresis pseudoinverse compensator based on the Preisach model [84]. The robustness of this control scheme to external disturbances has been demonstrated in experiments. With the DEA model structure represented as shown in Fig. 10(a), a feedforward inverse compensator was designed to eliminate the hysteresis and creep effects, and a model predictive controller was utilized in series with the feedforward compensator to tackle the system dynamics, model errors and disturbances [75] (Fig. 15(c)). In addition, an adaptive generalized predictive controller combining the merits of model predictive control and adaptive control was used to better cope with model uncertainties and disturbances, which was cascaded with a feedforward controller compensating for the hysteresis and quadratic input nonlinearity [83] (Fig. 15(d)). In these control approaches except for the one in [38], the feedforward controller and feedback controller are connected in series.

In the second type of feedforward-feedback control approach, the feedforward controllers compensate for both hysteresis nonlinearity and vibrational dynamics. The basic idea of this type of control approach is to first develop a feedforward controller based on the DEA model to compensate for the hysteresis, creep, vibrational dynamics and quadratic input nonlinearity, and then design a feedback controller to reduce the adverse impacts of modeling errors and disturbances on the tracking performance. Therefore, the feedforward controller and feedback controller are connected in parallel in this type of feedforward-feedback control approach (i.e. feedforward-feedback compound control), as depicted in Fig. 12(c). Based on this idea, some works on DEA control have been proposed that combine different DEA models and different feedback control techniques. With the DEA modeled as a cascade of a quadratic term and a linear dynamics, Koshiya et al. [112] used the inverse of the identified dynamic model as a feedforward controller, which was combined with an anti-windup PI feedback controller. A square root function was inverted between the compound controller and the plant so as to eliminate the input quadratic nonlinearity, as shown in Fig. 16(a). With the DEA modeled as shown in Fig. 10(e), Wang et al. [86] firstly designed a model-inverse-based feedforward controller to eliminate the quadratic nonlinearity, hysteresis and vibrational dynamics, and then employed the fuzzy logic PI controller to tackle the unknown disturbances (Fig. 16(b)). Different from using the analytical inverse of the DEA model in [86], Zhang et al. [95] directly modeled the inverse dynamics of DEAs by utilizing a GRU neural network. The data-driven inverse model was used as a feedforward controller and then combined with a PI controller, as shown in Fig. 16(c). Without the need for model inversion, Huang et al. [113] employed the iterative learning control method in conjunction with a physics-based dynamic model to obtain the feedforward control sequence, which was combined with a PI controller based on the self-sensing displacement feedback (Fig. 16(d)).

*3) Adaptive feedforward control:* The material parameters in DEAs are slowly time-varying over a long period of time due to the viscoelasticity and aging of elastomer materials. Moreover, there are model uncertainties and unknown disturbances for DEAs. Adaptive control has gradually attracted attention in the tracking control of DEAs in recent years owing to its advantage in tackling the systems with uncertainties and parameter variations. The core idea of adaptive control is to adjust the parameters or structure of the controller in real time to adapt to changes in the DEA model or disturbance.

Various adaptive control schemes for DEAs have been developed, and their control architectures are shown in Fig. 17. Using the model structure shown in Fig. 10(e), Zhang et al. [80] proposed a control scheme based on model reference adaptive control (MRAC). As shown in Fig. 17(a), a feedforward inverse compensator was firstly designed to mitigate the quadratic input nonlinearity and hysteresis nonlinearity. With the identified third-order linear model as the reference model, the MRAC method was employed to make the output of the DEA track the output of the linear model. The inverse of the linear model was then placed in front of the linear model so as to make the DEA output track the desired displacement. In addition, self-tuning control was also used for the tracking control of DEAs [76]. As illustrated in Fig. 17(b), the model structure shown in Fig. 10(c) was used and the hysteresis model parameters were identified online according to the least-mean-square (LMS) algorithm. The inverse of the DEA model was adopted as a feedforward controller, and the controller parameters were also adjusted when the model parameters were updated. The robustness of the proposed controller to actuator load variations was verified experimentally. Zhao and Wen [114] proposed an adaptive inverse control scheme with hysteresis compensation, in which the DEA was modeled as a cascade of static hysteresis nonlinearity and linear dynamics. As shown in Fig. 17(c), an adaptive inverse controller, cascaded with a direct inverse hysteresis compensator, was used to cope with the vibrational dynamics and uncertainties. Different from the adaptive controller in [76] adopting the inverse of the identified system model, this adaptive controller adopted a finite-impulse-response filter and its weights were updated in real time to minimize the tracking error. Considering that the secondary path (i.e. the path from the output of the adaptive controller to the error signal) causes phase shifts or delays in signal transmission, the filtered-x normalized LMS (NLMS) algorithm was used where the secondary path was estimated online. This control method achieved accurate tracking control of DEAs over a wide operating frequency range. Another adaptive feedforward control scheme is bioinspired adaptive control based on cerebellar model articulation controller (CMAC) [115]. A perfect controller including a model-inverse-based feedforward controller and an error feedback controller was designed, which was then approximated by a Gaussian basis function CMAC. The CMAC output was followed by a saturation function constraining the input to the DEA within an acceptable range and then a square root function to eliminate the quadratic nonlinearity, as illustrated in Fig. 17(d). The adaptive law of the weights in the CMAC neural network was derived by minimizing the tracking error cost function. This control scheme demonstrated satisfactory robustness against external disturbances in experiments.

The comparison of different control methods for soft DEAs in terms of highest tracking frequency, maximum travel range and robustness is presented in Table II. It can be seen that most control methods can only achieve precise trajectory tracking of DEAs at low frequencies or under low travel ranges (neglecting hysteresis nonlinearity).

## V. Self-sensing Methods

In the closed-loop control of DEAs, the displacement feedback signal is required. Using additional external sensors to provide displacement feedback is an effective means. This, however, not only increases costs, but is not feasible in many applications of DEAs, especially in unstructured, narrow and extreme environments. Utilizing the DEA itself to sense its own strain, also known as self-sensing, can fulfill the displacement detection requirement without the need for additional sensors. Hence a more compact closed-loop system can be achieved by using the displacement obtained through the self-sensing algorithm as feedback.

The self-sensing principle of DEAs is that the electrical parameters of a DEA changes with the DEA deformation. Thus, a common type of self-sensing approach is to first identify the electrical parameters by measuring the electrical quantities and using suitable identification algorithms, and then reconstruct the DEA displacement based on its relationship with electrical parameters (Fig. 18(a)). Applying a voltage to the DEA causes an expansion of the electrode area and a reduction in the thickness of the dielectric film, further leading to changes in the DEA capacitance, dielectric resistance and electrode resistance. Theoretically, all these electrical parameters can be used to estimate the DEA displacement. However, in fact, it is difficult to accurately estimate the dielectric resistance, because the dielectric resistance is typically in the megaohm range (resulting in a very small leakage current). The electrode resistance is dependent on many factors, including not only the DE strain but also the electrode thickness, electrode resistivity (varying with the number of cycles), as well as the electrode material [50], making it difficult to derive an exact expression relating the electrode resistance to the DEA deformation. Moreover, the experiments have shown that the electrode resistance exhibits a non-monotonic and obviously hysteretic dependence on the DEA displacement, while the DEA capacitance exhibits a monotonic and almost non-hysteretic dependence on the DEA displacement [54], [116], [117]. Therefore, capacitive self-sensing is the most suitable for DEAs.

There have been some works reconstructing the DEA displacement by utilizing the change in capacitance. As detailed in Section III-A-1, a DEA can be electrically modeled as a capacitance connected in parallel with a leakage resistance ($\sim G\Omega$) and in series with a lumped electrode resistance ($\sim M\Omega$), or a leakage-free RC series circuit, which acts as a high-pass filter. In light of this, the input voltage for performing simultaneous actuation and sensing in a DEA is usually the superposition of a low-frequency, high-amplitude actuation voltage and a high-frequency, low-amplitude sensing voltage [54], [55], [112], [116], [117], [118]. The low-frequency, high-amplitude voltage component results in insignificant current and therefore does not affect self-sensing; conversely, the high-frequency, low-amplitude voltage component does not contribute to actuation.

Some capacitance estimation methods have been developed. A typical estimation method is based on the electrical impedance analysis. With adopting the RC series equivalent circuit, Cheng et al. [118] utilized the amplitude ratio and phase difference between current and sinusoidal input voltage (i.e. the modulus and angle of the complex impedance) to calculate the DEA capacitance. Cao et al. [115] connected a known auxiliary resistor in series to the RC circuit and used two high-frequency low-amplitude voltage signals for self-sensing. Based on the sensing voltages and the voltage across this auxiliary series resistance, the DEA capacitance was calculated through impedance analysis. Another capacitance estimation method is online linear regression. Based on the full electrical model or leakage-free electrical model of DEAs, Professor Rizzello's research group [54], [112], [116], [117], [119] derived continuous-time circuit equations containing the derivative of the charge on the capacitor. By discretizing the circuit equations and assuming that the changes in electrical parameters between two successive samples are much smaller than the changes in the DEA voltage and current, a difference equation with linear-in-parameters structure was obtained. By using the forgetting factor recursive least squares (FFRLS) algorithm, the DEA capacitance could be estimated online from the real-time measurements of voltage and current. It should be pointed out that the above capacitive self-sensing approaches for DEAs require additional high-frequency sensing voltage superposed to the actuation voltage. To overcome the limitation, a capacitance estimation method based on the extended Kalman filter (EKF) was proposed [120]. Specifically, based on the RC parallel-series equivalent circuit and electromechanical coupling modeling of DEAs, a nonlinear state-space model with the capacitance and the charge on the capacitor as state variables, the voltage applied to the DEA as output variable and the resulting current as input variable was developed. The DEA capacitance was then estimated in real time by means of an EKF. This method can operate with any arbitrary driving voltage, without requiring the addition of particular sensing voltage. Once the DEA capacitance estimate is obtained, the DEA displacement can be reconstructed by utilizing the relationship between the DEA's capacitance and displacement. The relationship can be determined through polynomial fitting [54], [116], [117], piecewise linear fitting [118], using an autoregressive exogenous model [115], or physical derivation (which may involve the capacitance formula of plate capacitors as well as the deformation mode and geometry of DEAs) [55], [120], [121].

Another type of self-sensing approach for DEAs is the data-driven approach, which directly establishes the relationship between the DEA's electrical quantities and displacement (Fig. 18(b)). Different from the capacitance-based self-sensing approach, the data-driven self-sensing approach does not require an electrical model and does not need to estimate electrical parameters. Ye et al. [122] employed the second-order polynomial fitting, artificial neural network (a two-layer neural network trained through back-propagation algorithm), enhanced artificial neural network with data pre-processing, and nonlinear autoregressive neural network with exogenous inputs (NARX neural network) to model the nonlinear relationship between the magnitude of sensing signal's current and the DEA displacement. Among these, the NARX neural network achieved the lowest displacement estimation error. Huang et al. [113] also utilized a NARX neural network to establish a displacement self-sensing model for DEAs, in which the model inputs were the driving voltage, current, and time integral of current. A detailed comparison of different self-sensing methods for soft DEAs is given in Table III.

There have already been some successful attempts to achieve sensorless control by using the displacement reconstructed by self-sensing as a feedback signal in the closed-loop operation [54], [55], [112], [113], [115], [121]. The performance of closed-loop control based on self-sensing feedback and that based on sensor feedback have also been experimentally compared [54], [112], [113]. The results show that the overall performance of the self-sensing based control is satisfactory, but it is strictly speaking not as good as the performance of the sensor-based control. In addition, the decline in control performance caused by self-sensing is almost negligible in the case of slow-varying reference signals, but it shows an increasing trend as the controller gain or the reference signal frequency increases. The reason behind the performance degradation is twofold: the self-sensing tends to amplify measurement noise, and the self-sensing introduces delay in the feedback loop, which may eventually produce destabilizing effects.

## VI. Summary and Outlook

DEAs have received a lot of attention and extensive efforts have been made in this research field over the past few decades. However, there is still a long way to go before the DEAs can be truly applied in practice. Precise control is a major challenge limiting the widespread use of DEAs, although there has been less focus on the control of DEAs compared to the development of new DE materials, new electrode materials, improved fabrication techniques and new actuator configurations to provide better actuation performance. Additionally, modeling is critical to understand the electromechanical behaviors of DEAs and develop model-based controllers. Sensing is also of critical importance for achieving precise closed-loop control. The self-sensing capability of DEAs can be utilized to provide feedback, thereby eliminating the need for additional external sensors in the closed-loop control of DEAs. The current research and future prospects of the modeling, control and self-sensing of DEAs are summarized as follows.

### A. Modeling

Both physics-based modeling and phenomenological modeling of DEAs go through a period from static modeling to comprehensive dynamic modeling. Physics-based models can provide physical insights into various electromechanical behaviors, but satisfactory model accuracy is generally achieved within a small travel range or at low frequencies. Phenomenological models describe the behaviors of DEAs based on phenomenological characteristics and are more commonly used in model-based control. The hysteresis behavior is particularly difficult to describe using mathematical models, because the hysteresis nonlinearity of DEAs is both amplitude-dependent and rate-dependent. Various hysteresis models have been proposed, among which the P-I model and its variations are the most popular, and there are also a few attempts to model the hysteresis using gated recurrent neural network. However, developing a model that can accurately describe the hysteresis of DEAs over the entire operating frequency range and constructing the hysteresis inverse remain challenging.

### B. Control

There are increasing research efforts in the area of precise control of soft DEAs. Early efforts have mostly used model inversion-based feedforward control and feedback control. Later, feedforward-feedback control methods have been mostly used. In these feedforward-feedback control methods, the feedforward controllers are designed for compensating the hysteresis, creep, quadratic input nonlinearity and/or vibrational dynamics, and the feedback controllers are designed to deal with compensation errors caused by model inaccuracy or approximate inverse compensation and external disturbances. Considering the fact that the DEAs exhibit time-dependent viscoelastic behavior and may have poor property consistency and wide manufacturing tolerances, adaptive control strategies are becoming promising in recent years. Data-driven methods including neural networks and reinforcement learning for DEA control have been gradually recognized. It is important to note that the existing control methods for DEAs are generally limited to the cases of low frequencies or small travel ranges, and few control methods can achieve high-precision tracking control of DEAs at high frequencies and large strokes.

### C. Self-sensing

The self-sensing methods for DEAs can be classified into two categories: model-based self-sensing methods and data-driven self-sensing methods. In the model-based self-sensing methods, an equivalent circuit model of DEAs needs to be established. The DEA capacitance is estimated from the measurements of voltage and current, and then the DEA displacement is determined using the capacitance-displacement relationship. In the data-driven self-sensing methods, the relationship model between the DEA's electrical quantities and displacement is directly established using the collected experimental data. Some efforts have been made to achieve sensorless control by leveraging the self-sensing capability of DEAs. However, due to the delay and noise amplification introduced by the self-sensing, the use of self-sensing feedback leads to a certain degree of degradation in control performance compared with sensor-based feedback, especially in cases of high frequency and high controller gain. Thus, improving the self-sensing accuracy and self-sensing based closed-loop performance at high frequency needs to be addressed.

### D. Challenges and Perspectives

In summary, the current challenges in the modeling, control and self-sensing of soft DEAs include: (1) accurate modeling and inverse compensation of highly rate-dependent hysteresis over the entire operating frequency range; (2) high-bandwidth tracking control of DEAs; and (3) precise self-sensing of high-frequency displacement. Moreover, the existing research efforts in the modeling, control and self-sensing are mainly for single-degree-of-freedom (1-DOF) DEAs in simple configurations. However, in many cases, multi-degree-of-freedom (multi-DOF) DEAs are required to enable soft robots to perform complex tasks. For multi-DOF DEAs, the coupling effects between different DOFs need to be additionally taken into consideration in the modeling and control, as they can degrade the tracking accuracy. The control of DEAs and the soft robots driven by them remains an area worthy of further exploration. It is hoped that high-bandwidth sensorless control approaches for multi-DOF DEAs can be developed in the near future to promote the practical applications of soft DEAs.

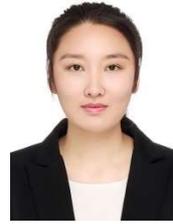
**Yunhua Zhao** received the B.S. degree in mechanical engineering from Chongqing University, Chongqing, China, in 2016, and the Ph.D. degree in mechanical engineering from Shanghai Jiao Tong University, Shanghai, China, in 2021.

She is currently an assistant professor at the School of Mechanical Engineering, Hebei University of Technology. Her current research interests include smart materials and structures, modeling and control of soft actuators and soft robots.

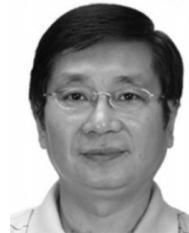
**Guang Meng** received the Ph.D. degree from Northwestern Polytechnical University, Xi'an, China, in 1988. From 1989 to 1993, he was a Research Assistant with Texas A&M University, College Station, an Alexander von Humboldt Fellow with Technical University Berlin, Berlin, Germany, and a Research Fellow with New South Wales University, Sydney, Australia.

He is currently a Chair Professor of the State Key Laboratory of Mechanical System and Vibration, School of Mechanical Engineering, Shanghai Jiao Tong University. His research interests include smart materials and structures, nonlinear dynamics and vibration control, and microelectromechanical systems.